%% file: root.tex
\documentclass[letterpaper, 10 pt, conference]{ieeeconf}  % Comment this line out if you need a4paper

\IEEEoverridecommandlockouts                              % This command is only needed if 
\usepackage{amsmath} % assumes amsmath package installed
\usepackage{amssymb}  % assumes amsmath package installed
\usepackage{siunitx}
\usepackage{multirow}
\usepackage{xspace}

\usepackage{tikz}
\usepackage{pgfplots}
\usepackage{tikzscale}
\usepackage{marvosym}
\usepackage{balance}
\usepackage{hyperref}

\pgfplotsset{compat=1.18}
\usetikzlibrary{external}
\usetikzlibrary{matrix, decorations.pathreplacing, calc, positioning, fit, fadings, shapes.arrows, arrows.meta, shapes.geometric, backgrounds}
\usetikzlibrary{shadings}
\usetikzlibrary{shapes.arrows}
\definecolor{custom-blue}{HTML}{1a80bb}
\definecolor{custom-red}{HTML}{a00000}
\definecolor{custom-yellow}{HTML}{f2c45f}
\definecolor{muted-gray}{HTML}{808080}
\definecolor{muted-gold}{HTML}{f0c571}
\definecolor{muted-teal}{HTML}{59a89c}
\definecolor{muted-blue}{HTML}{0b81a2}
\definecolor{muted-red}{HTML}{e25759}
\definecolor{muted-darkred}{HTML}{9d2c00}
\definecolor{muted-purple}{HTML}{7e4794}
\definecolor{muted-green}{HTML}{36b700}

\makeatletter
\pgfdeclareshape{motshape}{
    \inheritsavedanchors[from=rectangle] % basic rectangle shape
    \inheritanchorborder[from=rectangle]
    \inheritanchor[from=rectangle]{center}
    \inheritanchor[from=rectangle]{north}
    \inheritanchor[from=rectangle]{south}
    \inheritanchor[from=rectangle]{west}
    \inheritanchor[from=rectangle]{east}
    \inheritanchor[from=rectangle]{north west}
    \inheritanchor[from=rectangle]{north east}
    \inheritanchor[from=rectangle]{south west}
    \inheritanchor[from=rectangle]{south east}

    \backgroundpath{
        \southwest \pgf@xa=\pgf@x \pgf@ya=\pgf@y
        \northeast \pgf@xb=\pgf@x \pgf@yb=\pgf@y

        \pgfmathsetlengthmacro{\w}{\pgf@xb-\pgf@xa}
        \pgfmathsetlengthmacro{\h}{\pgf@yb-\pgf@ya}

        \pgfmathsetlengthmacro{\notch}{0.2*\w}

        \pgfpathmoveto{\pgfpoint{\pgf@xa}{\pgf@yb}}
        \pgfpathlineto{\pgfpoint{\pgf@xb-\notch}{\pgf@yb}}
        \pgfpathlineto{\pgfpoint{\pgf@xb}{0.5*\pgf@ya+0.5*\pgf@yb}}
        \pgfpathlineto{\pgfpoint{\pgf@xb-\notch}{\pgf@ya}}
        \pgfpathlineto{\pgfpoint{\pgf@xa}{\pgf@ya}}
        \pgfpathlineto{\pgfpoint{\pgf@xa+\notch}{0.5*\pgf@ya+0.5*\pgf@yb}}
        \pgfpathclose
    }
}
\makeatother

\newcommand{\ours}{MUFASA\xspace}
\newcommand\copyrighttext{\footnotesize \textcopyright~2025 IEEE. Personal use of this material is permitted. Permission from IEEE must be obtained for all other uses, in any current or future media, including reprinting/republishing this material for advertising or promotional purposes, creating new collective works, for resale or redistribution to servers or lists, or reuse of any copyrighted component of this work in other works.%
DOI: \href{https://ieeexplore.ieee.org/document/11423459}{10.1109/ITSC60802.2025.11423459}
}

\newcommand\copyrightnotice{%
    \begin{tikzpicture}[remember picture,overlay]%
     \node[%
        anchor=south, %
        yshift=10pt%
    ] at (current page.south)%
     {\fbox{\parbox{\dimexpr\textwidth-\fboxsep-\fboxrule\relax}{\copyrighttext}}};%
     \end{tikzpicture}%
}

\makeatletter
\newcommand\HUGE{\@setfontsize\Huge{40}{50}}
\makeatother

\newtheorem{definition}{Definition}

\title{\LARGE \bf
Multi-Staged Framework for Safety Analysis of Offloaded Services in Distributed Intelligent Transportation Systems % WIP
}

\author{Robin Dehler, Oliver Schumann, Jona Ruof, and Michael Buchholz% <-this % Check author list. 
\thanks{This work has been financially supported by the Federal Ministry of Education and Research (project autotech.agil, FKZ 01IS22088W). Parts of this work were supported by the State Ministry of Economic Affairs, Labour and Tourism Baden-Württemberg (project U-Shift\,II, AZ\,3-433.62-DLR/60).}%
\thanks{All authors are with the Institute of Measurement, Control and Microtechnology, Ulm University, Albert-Einstein-Allee 41, 89081 Ulm, Germany {\tt\footnotesize \{firstname\}.\{lastname\}@uni-ulm.de}}%
}

\newcommand{\ros}{ROS~2\xspace}

\begin{document}

\maketitle
\thispagestyle{empty}
\pagestyle{empty}

%%%%%%%%%%%%%%%%%%%%%%%%%%%%%%%%%%%%%%%%%%%%%%%%%%%%%%%%%%%%%%%%%%%%%%%%%%%%%%%%
\begin{abstract}
The integration of service-oriented architectures (SOA) with function offloading for distributed, intelligent transportation systems (ITS) offers the opportunity for connected autonomous vehicles (CAVs) to extend their locally available services.
One major goal of offloading a subset of functions in the processing chain of a CAV to remote devices is to reduce the overall computational complexity on the CAV.
The extension of using remote services, however, requires careful safety analysis, since the remotely created data are corrupted more easily, e.g., through an attacker on the remote device or by intercepting the wireless transmission.
To tackle this problem, we first analyze the concept of SOA for distributed environments.
From this, we derive a safety framework that validates the reliability of remote services and the data received locally.
Since it is possible for the autonomous driving task to offload multiple different services, we propose a specific multi-staged framework for safety analysis dependent on the service composition of local and remote services.
For efficiency reasons, we directly include the multi-staged framework for safety analysis in our service-oriented function offloading framework (SOFOF) that we have proposed in earlier work.
The evaluation compares the performance of the extended framework considering computational complexity, with energy savings being a major motivation for function offloading, and its capability to detect data from corrupted remote services.
\end{abstract}

\copyrightnotice

%%%%%%%%%%%%%%%%%%%%%%%%%%%%%%%%%%%%%%%%%%%%%%%%%%%%%%%%%%%%%%%%%%%%%%%%%%%%%%%%
\section{INTRODUCTION}
The integration of autonomous vehicles has the potential to provide major benefits for road transport and intelligent transportation systems (ITS), e.g., the reduction of accidents or traffic congestion.
However, when developing functions for autonomous driving, safety is of utmost importance.
If the trajectory that an autonomous car follows is not safe, accidents may occur with even fatal outcome, threatening not only the passengers of the autonomous vehicle, but also other traffic participants.
Thus, a careful safety analysis of autonomous driving functions is essential for further integration of autonomous cars in real-world traffic~\cite{liu19}.

Due to the complex nature of the autonomous driving task, it is considered that autonomous vehicles are not solely capable of efficiently solving all situations with their onboard computational hardware.
Thus, a lot of approaches nowadays consider distributed scenarios, where connected autonomous vehicles (CAVs) have the possibility to use different distributed functions in form of different services within a distributed service-oriented architecture (SOA)~\cite{kampmann19}.
Safety analysis is even more important, if we consider a distributed environment, since CAVs not only need to depend on their own capabilities, but also that of other connected traffic participants, including both other CAVs, and other remote devices, e.g., Multi-Access Edge Computing (MEC) servers~\cite{liu19}.
% For the sake of simplicity, we use the terms CAV and MEC server for all types of autonomous vehicles and remote devices in the remainder of this paper, respectively.

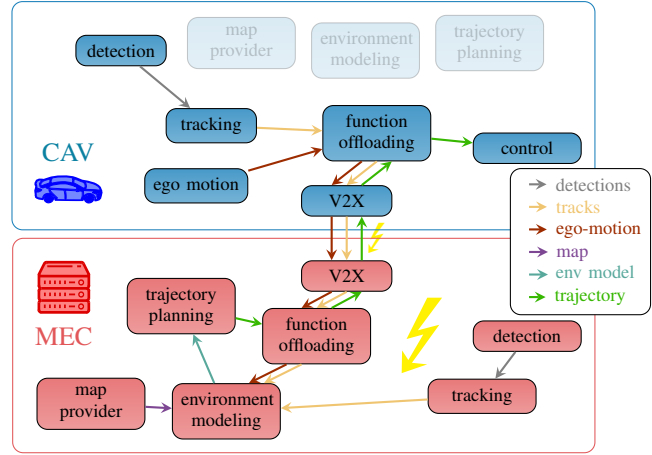
\begin{figure}[t]
    \centering
    \input{img/overview.tex}
    \caption{Simplified service configuration for autonomous driving. Deactivated nodes are more transparent. The legend on the right shows different arrow colors indicating data types that are sent through the interfaces. The lightnings show possible areas for data manipulation for remote services.} 
    \label{fig:intro}
\end{figure}

In distributed environments, function offloading offers the possibility to offload different components of the execution chain for the autonomous driving task to remote devices.
Function offloading has the potential to reduce energy consumption and the needed computational capacity of CAVs with limited resources~\cite{dehler25}.
For CAVs, offloaded functions may vary from components at the beginning of the processing chain, e.g., multi-object tracking (MOT) to the end, e.g., trajectory planning.
The data provided by a remote device needs to be handled even more carefully, since it is much more likely to be manipulated due to less controllability of the remote device and the wireless transmission medium~\cite{gao22}.
An exemplary service configuration considering local and remote services for the autonomous driving task of a CAV is seen in Fig.~\ref{fig:intro}, where the possible safety-critical areas, i.e., the wireless connection and the MEC server in general, are indicated with lightnings.
Consequently, when using offloaded functions, a local safety component is crucial.

In this paper, we analyze safety in distributed environments for CAVs and ITSs in general, considering a service-oriented architecture (SOA) in which the CAVs can use a function offloading framework to distribute different functions in form of services to remote devices.
For this, we derive a hierarchical structure for these services, dependent on the solved tasks and the location of execution, that are needed to handle the autonomous driving task.
Resulting from this analysis, we propose a \textbf{MU}lti-Staged \textbf{F}r\textbf{A}mework for \textbf{S}afety \textbf{A}nalysis (\ours) that we include in our service oriented function offloading framework (SOFOF) proposed in~\cite{dehler25}.
The extended framework considers both quality of service (QoS), which we have already done in~\cite{dehler25} as well as the data quality received by offloaded functions.
We evaluate different aspects of function offloading and the safety framework considering different service configurations and manipulated data in simulation, showing the safe handling of the autonomous driving task even if function offloading fails due to external failures.
%Furthermore, we apply our approach in our real-world test vehicle to show the applicability of the framework as a whole for real-world scenarios.

Our main contributions are as follows:
\begin{itemize}
    \item We extend the definitions for SOAs so that a different safety analysis for local and remote services is possible.
    \item We propose \ours that is applicable to different service configurations. We have integrated the proposed safety framework in our previously proposed SOFOF. 
    \item We evaluate \ours for different offloading scenarios with different attacks to the simulated MEC server and compare its performance to SOFOF without the safety framework. 
\end{itemize}

\section{Related Work}
With being an important topic for both the development as well as the integration of CAVs, there exists a large variety of research topics on safety for autonomous driving in general, SOAs, and distributed systems.
Research was also conducted to standardize safety requirements for autonomous driving. 
Note that we specifically only consider safety, and not security, which is out of scope for this work.
Nonetheless, even with multiple security stages between remote and local devices, attacks can never be completely excluded~\cite{liu19}. % TODO maybe a 2nd citation
Thus, given that security breaches are not excludable, the need for a local safety check will always be existent.

\subsection{Safety in Autonomous Driving}
Depending on the considered function, different approaches for safety are employed.
The term self-assessment is also used to describe the intrinsic monitoring of these components.
A subjective logic (SL) based approach to monitor MOT performance for a classical tracking algorithm was proposed in~\cite{griebel24}.
SL methods have also been applied to grid maps to assess conflicting sensor data for robust motion planning~\cite{schumann24}. 
For motion and trajectory planning in general, collision avoidance is an important safety aspect that has been studied profoundly~\cite{ziegler10,mirchevska18,huang23,genin23}. % TODO more citations
Recently, the end-to-end (E2E) learning paradigm has attracted significant interest in the context of autonomous driving.
A thorough survey about E2E learning and the importance of safety for E2E learning approaches is given in~\cite{chib_23}.

\subsection{Safety for Service-Oriented Architectures}
In~\cite{kampmann19}, the dynamic nature of SOAs for safety is discussed. The results obtained imply that, if designed properly, the orchestrator within a SOA can improve safety, offering the possibility to analyze all relevant interfaces given QoS requirements.
Specific QoS constraints can, e.g., be used to analyze the state of a service and change the service configuration, if the constraints are not satisfied, which we have already conducted in our previous work~\cite{dehler25}.
The compatibility of SOA with safety-critical applications was also investigated in~\cite{kugele17}.

\subsection{Safety for Distributed Systems}
For distributed systems, it is especially important to analyze the data received from other computational devices.
A thorough survey on how to guarantee safety for CAVs in edge computing systems was conducted in~\cite{liu19}.
Zheng~\textit{et al.} specifically analyze the benefits of distributed systems for autonomous driving, for both non-safety and safety relevant services~\cite{zheng15}.
A decentralized SOA approach for distributed environments, including safety consideration, was proposed in~\cite{schindewolf22}, with the goal to create a resilient SOA.

\subsection{Standardization}
To be able to better classify matters of safety, multiple standards have been developed.
The functional safety of road vehicles is addressed in ISO 26262~\cite{iso26262}.
In ISO 20077~\cite{iso20077}, the concept of extended vehicle is introduced including external communication, with a follow-up given in ISO 20078~\cite{iso20078}.
ISO 21217~\cite{iso21217} considers ITS communication networks and describes employment and communication modes for ITS stations in communication networks.

\section{Safety Analysis for Distributed Service-Oriented Architectures}
In this section, we derive a methodology for a safety analysis of distributed SOAs.
For this, we first derive our understanding for a service in the context of autonomous driving from~\cite{kampmann19}.
Further, we extend the service definition for distributed systems.
\begin{definition}[Service]\label{def:service}
Given $\mathbb{S}_t$ as a set of $\sigma_t$ available services $S_i$ at time step $t$, $1\leq i\leq\sigma_t$, $\sigma_t>0$, we define service $S_i\in\mathbb{S}_t$ with the tuple
\begin{equation}\label{eq:service}
    S_i=(R_i,G_i,l_i).
\end{equation}
Each service has its dedicated set of requirements $R_i=\emptyset$ or $R_i=\{r_i^1,\ldots,r_i^{\mu(i)}\}$, a set of fixed guarantees $G_i=\emptyset$ or $G_i=\{g_i^1,\ldots,g_i^{\kappa(i)}\}$, and an integer value $l_i$ indicating the enumerated available computing device, i.e., a station~ID.
\end{definition}
In Def.~\ref{def:service}, $R_i$ and $G_i$ are taken from~\cite{kampmann19}, while $l_i$ depicts the proposed extension for distributed environments. 
It is also worth mentioning that in contrast to the definition in~\cite{kampmann19}, we explicitly do not fix the number of available services $\sigma_t$, which is dependent on the availability of services from remote devices at time step $t$.
We use $\mathbb{S}^{(l_i)}$ as the set of available services at computing device $l_i$ and $\mathbb{S}$ as the set for all local services.
From this follows $\mathbb{S}\subseteq\mathbb{S}_t$ $\forall t$.

Generally, in SOAs, a service orchestrator handles the state transitions of services and manages the service configuration.
The task of the service orchestrator is to find a set of local or distributed services or both, from last to first, so that every requirement of each service is fulfilled, with the first service having no service requirement, e.g., for CAVs, a camera service guaranteeing images.
With the extended service definition, we define a switch in the computing device.
\begin{definition}[Device switch]
     Given the two services $S_j$ and $S_k$, $j,k<\sigma_t$, with compatible guarantees $G_j$ for requirements $R_k$, a device switch from service $S_j$ to $S_k$ is defined, if $l_j \neq l_k$.
\end{definition}
The orchestrator can monitor device switches between different services, which is necessary for different handling of local and remote services.
In~\cite{kampmann19}, the term \textit{service composition} is used to define a set of services with consecutive connected requirements and guarantees, usually with the goal to solve a specific task, e.g., vehicle actuation for autonomous driving.
In the following, we use the term \textit{distributed service composition} to describe service compositions with at least one device switch.

\begin{figure}[t]
    \centering
    \vspace{.2cm}
    \input{img/update_hierarchical_structure.tex}
    \caption{Exemplary hierarchical tree structure of services for a distributed ITS considering one CAV and a MEC server. The color indicates the location of the services, respectively. A distributed service composition can be chosen from the root nodes, i.e., CAV and MEC detection, to the leaf node CAV actuation.}
    \label{fig:hierarchy}
\end{figure}
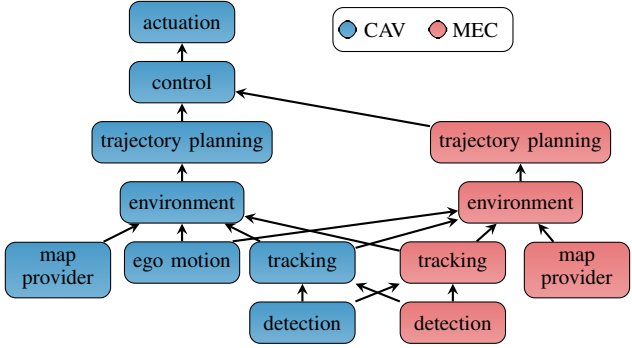

All possible configurations can be composed in a hierarchical graph, with the root node representing the last service of the execution chain, while each leaf node represents the first with no requirement.
Fig.~\ref{fig:hierarchy} shows an exemplary structure, from the bottom to the top, for one CAV at time step $\tau$ with $\mathbb{S}_\tau = \mathbb{S}\cup\mathbb{S}^{(1)}$, given locally available services $\mathbb{S}$ and remote services $\mathbb{S}^{(1)}$ from a MEC server with $l_i=1$.

In our previous work, we have introduced two operating modes of SOFOF, i.e., as a service provider offering and handling services from remote devices to the considered CAV, and as a service requester handling service offers and the local services onboard the vehicle~\cite{dehler25}.
%The service provider and service requester jointly operate as the service orchestrator for all available services on the corresponding CAV.
Both together operate as the service orchestrator for all available services on the corresponding CAV.
In Fig.~\ref{fig:intro}, these two are also visualized as the two function offloading services, where the service on the CAV operates as a service requester and the one on the MEC as a service provider.
It is also shown that all data that are exchanged between the CAV and the MEC are first passed through the function offloading services, before being sent through a vehicle-to-anything (V2X) service.

Given the extended service definition in Eq.~\eqref{eq:service}, the service requester has an overview of the services which are currently offloaded and which are local.
Thus, each composition of offloaded and local services can be handled differently concerning safety and QoS requirements.
Note that each service requester may also manage concurrent services from multiple service provider instances.
Equally, one service provider may offer services to multiple requesters simultaneously.

The motivation for this decentralized orchestration of services was that each computational unit maintains control over its own services, with the disadvantage that the CAV does not have direct control over the offloaded services.
However, a centralized orchestrator approach would infer a single point-of-failure, which would again dictate a safety analysis, including even the local services.

\section{Multi-staged Framework for Safety Analysis}
To ensure safe CAV operation, a safety framework must consider all possible service configurations and analyze the data received from remote services as well as QoS considerations.
The framework must also consider that data from preceding services might not be available for safety analysis as a consequence to offloading.
Note that in~\cite{dehler25}, we have already analyzed two QoS requirements that are already considered in SOFOF, i.e., maximum latency $l_\text{max}$ and maximum inter-arrival time $\Delta t_\text{max}$.
If at least one of these requirements is violated by a remote service, the offloading for this particular service is stopped and simultaneously the local service reactivated.

Our proposed \ours, however, analyzes the data quality.
It is deeply interconnected with SOFOF itself, to directly choose a particular safety stage dependent on the service composition.
The close interconnection guarantees a direct reaction to failures and manipulated data.
One of the main motivations for function offloading in general was to reduce energy consumption on the battery-powered CAVs.
Consequently, a major goal was to also create a simplistic, but efficient safety framework. 

\subsection{Considered Services for Offloading}
In our previous work, we have analyzed the application of SOFOF for the service trajectory planning.
In this work, we handle multiple services, i.e., multi-object tracking (MOT), environment modeling (ENV) and trajectory planning (TPL).
The three services MOT, ENV and TPL are handling typical subtasks for the overall task of autonomous driving.
The services are also seen in the hierarchical structure in Fig.~\ref{fig:hierarchy}.

The service MOT has the task to collect detections from multiple (redundant) detectors, e.g., camera, radar, or lidar, and create tracks of multiple objects including IDs and predictions. In our case, we use an extension of the labeled multi-Bernoulli (LMB) filter. The particular algorithm can be found in~\cite{aduulmttb2025}.
The service ENV takes the present tracks, the current ego motion information, and additional map data and creates a comprehensive environment model.
The environment model is then used by the TPL service to create trajectories. Our offloaded trajectory planning algorithm uses a Model Predictive Control (MPC) method with an Augmented Lagrangian-Iterative Linear Quadratic Regulator (AL-ILQR) optimizer to compute trajectories~\cite{ruof24}.
In our architecture, the services ENV and TPL use a shared memory interface for efficiency reasons.
Thus, the orchestrator needs to choose a configuration, such that both services run on the same device.
The requirements and guarantees of each service are summarized in Table~\ref{table:service-summary}.
\begin{table}
\centering
\vspace{.2cm}
\caption{Service Requirements and Guarantees}
\begin{tabular}{c | c | c}
    \hline
    Service & Requirements & Guarantees \\
    \hline
    MOT & detections & tracks \\
    % \textit{FUSION} & tracks & fused tracks \\
    ENV & tracks, map data, CAV ego motion & environment model \\
    TPL & environment model, vehicle metadata & trajectories \\
    \hline
\end{tabular}
\label{table:service-summary}
\end{table}

Note that there are several other services that need to be considered for the task of autonomous driving, e.g., object detection service, vehicle control service, map service, and metadata service.
These are not considered for offloading due to the following reasons.
Object detection is done on video, radar, and lidar data, which, when offloaded, would require transmitting huge amounts of raw sensor data.
Consequently, offloading in the sense of transmitting locally recorded sensor data to a server for processing poses substantial challenges.
However, an alternative approach could involve to use infrastructure sensors and restrict the remote detection service to areas covered by these sensors.
For offloading the task MOT, e.g., we already make use of infrastructure sensors to create a more comprehensive environment model.
Whether it is possible to yet offload the detection task will be considered in our future work.
Since vehicle control is the last service for autonomous driving, which sends steering and acceleration commands directly to the vehicle interface, we argue that this task should always be processed locally.
For other services like metadata or map, we use a simple lookup of the data that are stored on the remote devices.

Given these requirements and assumptions, we have the following possible configurations of our local and offloaded services on the CAV:
\begin{itemize}
    \item MOT offloaded, ENV and TPL local
    \item MOT offloaded, ENV and TPL offloaded
    \item MOT local, ENV and TPL offloaded
\end{itemize}

\begin{figure}[t]
    \centering
    \vspace{.2cm}
    \input{img/framework.tex}
    % \vspace{-.5cm}
    \caption{Proposed \ours for different service configurations (blue=local, red=remote). If a validation block is green for a specific configuration it is always applied while yellow ones are only applied under certain conditions, e.g., if previous ones cannot be used.}
    %\vspace{-.5cm}
    \label{fig:framework}
\end{figure}
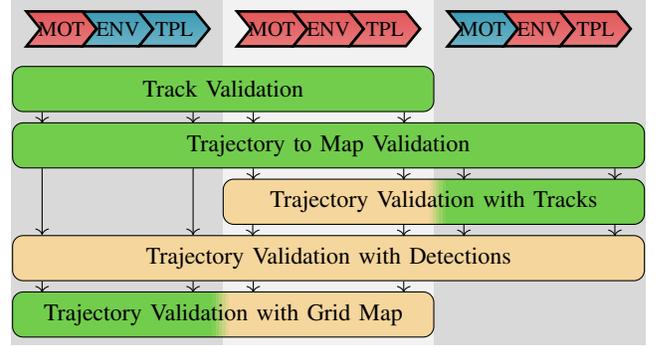

When being offloaded, we consider the corresponding local task to be deactivated.
Because of the different possible configurations, we propose a framework where multiple stages check the data received from offloaded services and, if necessary, also the data of concurrent services.
Fig.~\ref{fig:framework} shows the different validation checks that may be applied dependent on the service configuration, with green blocks indicating that this stage is always applied, and yellow blocks indicating that these are only used dependent on previous safety checks.
The application of multiple stages to different configurations is especially suitable in the context of SOAs, from which we motivate the concurrent usage of services.

The safety stages are sequentially applied from the top to the bottom.
Thus, if an earlier stage fails, there is no need to execute a later stage.
When a safety issue is detected, the implemented fallback strategy directly terminates all remote services and simultaneously activates the respective local ones for a safe execution.
Since for safe autonomous driving, the local functions need to be able to handle all possible driving scenarios, the fallback to only local execution is assumed to be the safest service configuration possible.
To also not fail directly again afterwards, a timer of time $t_\text{wait}$ is started, enforcing a waiting time before offloading is again possible.
The different safety stages and the dependencies are explained in the following.

\subsection{Track Validation}
To validate tracks, two different types of errors need to be considered, i.e., false positives and false negatives.
False negative tracks mean that real objects are not included in the respective list of tracks, while false positives imply that too many tracks are created or received.
The track errors can be recognized using locally created detections.
However, the difficulties in the validation of received tracks from a MEC server result from the fact that the field-of-view (FOV) of the local sensors may differ from the server and that detectors usually generate clutter measurements.

Thus, in a first step, the FOV of the CAV is determined with the Euclidean distance of the position of the CAV to the farthest detection.
Then, false positives can be detected by first arranging the Euclidean distances of all received tracks and the detections in an assignment matrix and then using the Hungarian method~\cite{kuhn55} to find a best match of tracks with detections.
A threshold value $\theta_\text{tr}$ is used to determine whether the average assignment cost of one track with a detection is too high.
This condition is summarized with
\begin{align}
    \frac{1}{n}\sum_{i=1}^n c_i < \theta_\text{tr},
\end{align}
with $c_i$ as the chosen assignment cost for a track $i$.

Using the detections to search for false negatives, however, is not quite trivial.
The problem here results from the clutter detections.
If a detection has no received track, it is not possible to say if it is missing or if the detection is just clutter.
A sort of tracking would be needed to use detections for validating false negatives, reducing the advantages of function offloading.
Thus, we do not directly check for false negative tracks, but use the detections and a different environment representation, i.e., a grid map\footnote{We did not yet consider a grid-map service in Figs.~\ref{fig:intro}, and~\ref{fig:hierarchy} and the previous descriptions, where we have an object-based environment representation. This is because the grid map is not directly utilized in our service composition for autonomous driving. However, having another approach of modeling the environment is beneficial for additional safety analysis, for both local and remote services, as it is explained in Secs.~\ref{tr-val-grid}.} to validate the subsequent service of trajectory planning, respectively.
If a false negative error is present in the track list, the trajectory would not avoid the corresponding object.
This results in the trajectory colliding with either the related detection or passing occupied cells in the grid map, or both.
If a possible collision is detected, the offloading of the MOT service is canceled and the local MOT service activated again.

\subsection{Trajectory to Map Validation}
As a first step for validating a trajectory, the trajectory is matched with the local map.
The matching is a straightforward comparison of the trajectory points of the whole trajectory with the route of the map.
The route contains points for a reference path as well as corresponding path boundaries.
% For efficiency reasons, we do not validate the whole trajectory, but only the points within a safety distance that is calculated by
% \begin{equation}\label{eq:safety-dist}
%     s_\text{CAV} = v_\text{now} \cdot \SI{1.8}{\second},  % Change this, it is better to match the whole trajectory
% \end{equation}
% with $v_\text{now}$ being the current velocity.
% The time $\SI{1.8}{\second}$ is used as safety distance, which then is half of the numerical value of the current velocity in $\text{km}/\text{h}$, which is also the legally required safety distance~\cite{bußgeld2025}.
If the Euclidean distance of a trajectory point to the reference path is greater a threshold $\theta_{map}$, the trajectory point is compared to the path boundaries.
If the point is outside the path boundaries, the map check fails and the function offloading canceled.

After the map validation is passed, the trajectory is further validated.
In this work, we perform an object-based trajectory validation, with tracks or detections, and a grid-based trajectory validation depending on the available data.

\subsection{Trajectory Validation with Tracks}
If there is a current track list present, the trajectory can be validated with the tracks, using a low-level collision check (LLCC).
For efficiency reasons, this check involves two steps.
First, the position of each track is compared to the position of the vehicle.
If, again, the Euclidean distance is smaller than a threshold $\theta_{obj}$, a more specific check involving the pose, i.e., position and heading, and the dimensions of the CAV and the object are compared.
If the two bounding boxes collide, the check evidently fails.
These two checks are done for each trajectory point that is within  a safety distance
\begin{equation}\label{eq:safety-dist}
    s_\text{CAV} = v_\text{now} \cdot \SI{1.8}{\second}
\end{equation}
with $v_\text{now}$ being the current velocity.
The time $\SI{1.8}{\second}$ is used as safety distance, which is half of the numerical value of the current velocity in $\text{km}/\text{h}$, which is also the mandatory safety distance in Germany~\cite{bußgeld2025}.
The future track states are predicted using a simple constant velocity (CV) model, again for efficiency reasons.

The early isolation of tracks farther away ensures that the more costly pose comparison is only done when necessary.
The second step proved to be effective, since using a fixed threshold would, e.g., be violated by oncoming cars even if the respective poses would not collide.

\subsection{Trajectory Validation with Detections}\label{tr-val-det}
If no current track list is available, or if the track list was determined to contain false data in the \textit{track validation} step, the trajectory is validated with the detections.
For this, all detections are first collected into one detection list.
Then, the same LLCC is applied to the detection list as before in the step \textit{trajectory validation with tracks}.
However, detections do not contain velocity information.
Thus, a prediction using, e.g., the CV model, is not possible.
Our goal was to develop a simple and efficient safety framework.
For that we consider static objects and compare the current position of each detection with each trajectory point until the emergency braking distance
\begin{align}\label{eq:brake-dist}
    s_\text{b} = \frac{v_\text{now}^2}{2\cdot a_\text{b}},
\end{align}
with $a_b$ being the absolute value of the maximum braking acceleration.

Moreover, if the CAV detects a leading vehicle, the distance to the leading vehicle must always be larger than a following safety distance $s_\text{follow}$, which is dependent on the current speed of the CAV $v_\text{now}$ and the leading vehicle $v_\text{lead}$.
The current speed of the leading vehicle can, e.g., be detected using front radar sensors.
The following safety distance is calculated by the difference of the braking distances of the CAV $s_\text{b,CAV}$ and the vehicle in front $s_\text{b,lead}$ plus the legally required safety distance $s_\text{CAV}$, see Eq.~\eqref{eq:safety-dist}. 
% It is legally required and is defined by $s_\text{r} = v \cdot \qty{1.8}{\second}$.
Hence, the necessary safety distance to a leading vehicle is defined by
\begin{align}
s_\text{follow} &= \max(s_\text{b,CAV} - s_\text{b,lead}, 0) + s_\text{CAV}.
\end{align}
Summarized, for a static environment, the trajectory is validated for a distance
\begin{align}\label{eq:static-valid}
    s_\text{valid} = \begin{cases}
        s_\text{follow}, &\text{if follow-up traffic} \\
        s_\text{b,CAV}, &\text{else}
    \end{cases}
\end{align}

As indicated in Fig.~\ref{fig:framework}, the \textit{trajectory validation with detections} is not always applied.
It is additionally used if other checks are not applicable, making it even more desirable to use a more conservative approach.
If this stage fails, lastly, the \textit{trajectory validation with a grid map} is applied.

\subsection{Trajectory Validation with a Grid Map}\label{tr-val-grid}
Depending on the service configuration, it is possible that the object-based trajectory validation, i.e., with tracks or detections, is not possible or insufficient to validate the trajectories.
Thus, to ensure the safety of the CAV, as a last stage, each planned trajectory is validated in a grid-based environment model.

A preliminary for the grid-based method is the existence of an occupancy grid map.
It divides the current environment into a large number of 2D cells.
Each cell models the state of occupancy, which is usually derived from lidar or radar measurements.
In this work, we use the grid-mapping approach presented in~\cite{wodtko23}.
Then, this grid is used in the sparse collision checking method proposed by~\cite{ziegler10}, in which an efficient collision checking method for vehicles in occupancy grid maps is proposed:
At first, the footprint of the vehicle whose trajectory is being validated is represented by several disks with identical radii.
Then, the grid-based environment is dilated by this radius.
At last, the poses of the trajectory are checked for collisions in the dilated grid by verifying that all disk centers lie within non-occupied cells.

Similar to the \textit{trajectory validation with detections}, the grid-map approach also only provides a static representation of the current environment.
No dedicated prediction of the moving objects can be drawn from them, nor is it possible to detect a leading vehicle.
Thus, in the \textit{trajectory validation with a grid map}, the trajectory needs to be validated within a distance $s_\text{b}$, see Eq.~\eqref{eq:brake-dist}.
If a possible collision is detected, the trajectory is labeled unsafe and the function offloading is canceled.

\section{Evaluation}
In this section, we evaluate the application of SOFOF with the extension of \ours.
In~\cite{dehler25}, we have introduced SOFOF as a \ros~\cite{ros2} node and used the concept of \ros lifecycle nodes for dynamic node activation and deactivation.
The general setup is the same for this evaluation, with the functionality of \ours included in the service requester, i.e., the considered CAV.

SOFOF itself already considers two QoS requirements.
The first requirement analyzes the latency $l$ between data generation on the remote device and the receiving on the local device, while the second analyzes the inter-arrival time $\Delta t$ between two consecutively received data elements.
In~\cite{dehler25}, we have determined the maximum values of the requirements for the task of TPL, which we now use for all considered services MOT, ENV, and TPL:
\begin{align}\label{qos-requirements}
\begin{aligned}
    l_\text{max}&=50\,\unit{ms} \\
    \Delta t_\text{max}&=100\,\unit{ms}\,.
\end{aligned}
\end{align}

With energy reduction being the major motivation for function offloading in general, we first evaluate the computational overhead of the extended SOFOF compared to the offloaded applications MOT, ENV, and TPL in simulation.
Then, we compare the performance of SOFOF with and without \ours by intentionally manipulating the data of offloaded services in simulation.

In~\cite{dehler25}, we have proposed two decision making algorithms that are used to decide when and where function offloading is available.
In the centralized offloading decision making (CODM), an offloading area is designated to each remote service.
We simulate multiple different traffic scenarios with multiple simulated cars and one simulated autonomous car that is considered as the ego vehicle.
The simulation environment is similar to a real world traffic layout located in Ulm-Lehr.
For the CODM, we chose different areas for the availability of MOT, ENV, and TPL remote services.
Note that for offloading the service MOT, we chose the area to be within the coverage area of simulated infrastructure sensors, such that the remote MOT service has the advantage of an extended environment model.
However, a remote MOT service could also be used with only local detections from a CAV that are sent to the MEC server.

The simulations are done in a distributed setup with two PCs, where on the first, the simulation including the simulated CAV and other simulated vehicles are running, and the MEC server including its functionality is simulated on the second.
The two PCs are connected through a WiFi network and an AMQP connection, with AMQP clients on both devices and the broker running on the second.
For an introduction to the AMQP protocol, we refer to~\cite{amqp12}.
The distributed simulation setup is used to emulate the connection for V2X communication.
For V2X communication in particular, we use interfaces standardized by the European Telecommunications Standards Institute (ETSI), i.e., Cooperative Awareness Message (CAM)~\cite{etsi_en_302_637-2_intelligent_2014}, Collective Perception Message (CPM)~\cite{etsi_tr_103_562_intelligent_2019}, and a custom Maneuver Coordination Message (MCM)~\cite{mertens21}\footnote{Currently, there is not yet a standardized version of the MCM available.}.
After careful tuning of the parameters of \ours, we used the parameter setup shown in Table~\ref{table:params} for the simulations\footnote{For real-world applications, it would be more useful to choose a longer wait time. However, to be able to evaluate more attacks, we intentionally selected a comparably short $t_\text{wait}$.}.
% TODO better explain setup and params

\begin{table}
\centering
\vspace{.2cm}
\caption{Simulation parameters}
\begin{tabular}{c | c | c}
    \hline
    Name & Sign & Value \\
    \hline
    Average assignment cost threshold & $\theta_\text{tr}$ & $\SI{0.2}{\meter}$ \\
    Trajectory to map threshold & $\theta_\text{map}$ & $\SI{0.3}{\meter}$ \\
    Object bounding box check threshold & $\theta_\text{obj}$ & $\SI{5}{\meter}$ \\
    Brake acceleration & $a_\text{b}$ & $\SI{8}{\meter\per\second\squared}$ \\
    Wait timer after failure & $t_\text{wait}$ & $\SI{10}{\second}$ \\
    \hline
\end{tabular}
\label{table:params}
\end{table}

\subsection{Computational Complexity}
If the computational complexity of the function offloading framework is too high, the main benefit of using offloaded services is annihilated.
As an indicator for computational complexity, we analyze the average CPU usage of SOFOF including \ours and SOFOF alone, i.e., without safety analysis, compared to the local services that may be offloaded.
For this, we simulate two similar scenarios, with the only difference being that \ours is included or not.
The simulation is running on an AMD Ryzen 9 7950x CPU.
Table~\ref{table:cpu-usage} shows the measured CPU usage in percent of one CPU core as well as the time how long the respective services were offloaded.
The upper and lower part of Table~\ref{table:cpu-usage} show the data for SOFOF+\ours and SOFOF alone, respectively. 
It can be seen that, when deactivating the local services, great savings concerning CPU usage are possible in both cases.
However, both SOFOF and SOFOF+\ours only add small CPU usages of $2.13\%$ and $2.3\%$, respectively.
Note that we do not manage the lifecycle state of SOFOF itself, which is why there is no data for the inactive state.
Also, the sum of the active and inactive time might not directly match with the total active time of SOFOF+\ours, which results from the needed time for state transitions.

Another insight that Table~\ref{table:cpu-usage} gives is that the usage of \ours reduces the offloading duration, especially for the services TPL and ENV.
This behavior is expected, and a consequence to \ours detecting possible critical situations and deactivating the function offloading.
The reduction of the total offloading time to almost the half for the TPL and the ENV service also result from the time $t_\text{wait}$, which the framework is waiting before the next offloading request.
Thus, a trade-off between safety and efficiency needs to be considered.

\begin{table}  %% Values taken from 20250430_084824_MUFASA
\centering
\caption{CPU Usage of Local Services}
\begin{tabular}{c | c | c}
    \hline
    \multirow{2}{*}{Service} & \multicolumn{2}{c}{CPU Usage $[\%]$ / Time $[s]$} \\
    & Active & Inactive \\
    \hline
    SOFOF+\ours & $2.3$ / $594.5$ & - \\
    MOT & $5.23$ / $551.6$ & $1.3$ / $42.1$ \\
    ENV & $18.75$ / $536.3$ & $1.2$ / $57.4$ \\
    TPL & $18.7$ / $537.6$ & $0.9$ / $54.9$ \\
    \hline
    SOFOF & $2.13$ / $594.5$ & - \\ % TODO
    MOT & $5.36$ / $541.0$ & $1.1$ / $53.3$ \\
    ENV & $18.6$ / $488.4$ & $0.7$ / $105.1$ \\
    TPL & $18.6$ / $488.9$ & $0.6$ / $105.1$ \\
    \hline
\end{tabular}
\label{table:cpu-usage}
\end{table}

\subsection{Service Manipulation}
To analyze the performance of SOFOF+\ours concerning the handling of corrupted or manipulated data, we manipulate the data that is received from local services using various attacks from an additional \ros node that exploits the node interfaces.
The attacker node first listens to available subscriptions and \ros services within the \ros framework.
The first attack directly corrupts the remote~TPL service by changing the map from the one set through the corresponding CAV to a different one, through a \ros service call.
When changing the map, the trajectory planner calculates trajectories for a different reference path, which in this case results in the trajectory leaving the corresponding lane and desired route of the CAV.
The resulting wrong trajectory that is sent to the corresponding CAV should be recognized by the \textit{trajectory to map validation} stage.
In a second attack, the attacker node first collects information about the interfaces over which the track lists are sent to the CAV and on the MEC server internally by listening to the published track lists.
Then, the attacker node spams the interface with empty track lists with a frequency of $\SI{100}{\hertz}$ that are used by the ENV~service on the MEC server or the CAV.
In both cases, one of the trajectory validation stages should detect a possible collision that results from planning trajectories with empty track lists.
Lastly, in a third attack, the track lists are manipulated such that false positive tracks are inserted, i.e., ghost objects, which should be detected by the \textit{track validation} stage.

\begin{figure}[t]
    \centering
    \vspace{.2cm}
    \includegraphics[width=.99\linewidth]{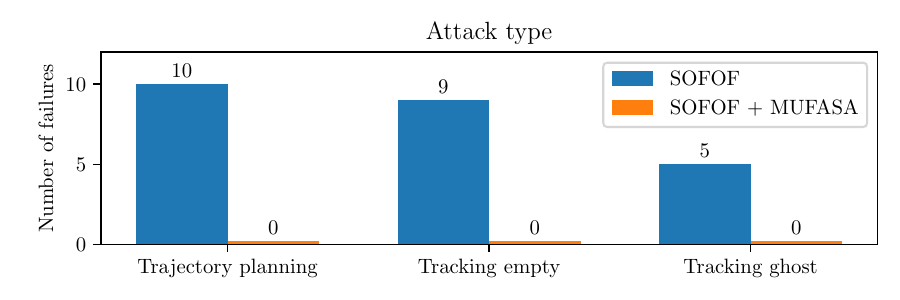}
    \caption{Number of failures during function offloading for SOFOF and SOFOF + \ours. For each attack, we have simulated $10$ scenarios. A failure is occurring if a collision happens or if a specified target zone is not reached within a time limit.}
    \label{fig:failures}
\end{figure}

We first analyze the effect of the attacks individually for the possible service configurations, comparing the performance of SOFOF alone to the extended SOFOF with \ours.
We simulate ten scenarios on two different routes in our simulation environment for each attack.
Fig.~\ref{fig:failures} shows a summary of how many failures occurred during function offloading in the simulations, grouped to the different attacks \textit{trajectory planning}, where the map of the TPL service is changed, and \textit{tracking empty} and \textit{tracking ghost}, where empty track lists and track lists with ghost objects are published, respectively.
The plot shows that SOFOF alone fails to detect manipulated data, since it does not analyze the data received from remote services.
A qualitative analysis shows that when spamming empty track lists, the CAV is mostly not able to stop behind a leading vehicle, if present.
Also, the changing of the map of the trajectory planning results in the vehicle leaving the corresponding routes and streets which also leads to multiple failures.
However, by extending SOFOF with \ours, the CAV is able to detect manipulated data and associated failures, enabling it to terminate function offloading early enough before any error occurs.
This enhances the overall safety of SOFOF.

Finally, we simulate four scenarios, each with one CAV and up to 30 other vehicles using the intelligent driver model (IDM)~\cite{treiber00}.
During the simulation, the attacker node randomly attacks active remote services with all three attacks that we have already investigated. %, and once without any attacks.
To be able to sample more attacks more frequently, we further reduced the wait time $t_\text{wait}$ to $\SI{2}{\second}$.
Table~\ref{table:two-cavs} summarizes for all scenarios how often each stage of \ours is triggered and how often a safety violation is detected and, thus, a fallback is triggered.
It can be seen in the table that throughout the four scenarios, each stage is triggered multiple times.
Note that stage triggering depends on the current offloading composition, as shown in Fig.~\ref{fig:framework}, and later stages are only triggered if no issue was detected in earlier ones.
If the MOT task is offloaded, stage (i) is always triggered when a track is received on the CAV.
Stage (ii) is triggered, when a trajectory is received while offloading is active.
Thus, the number of executions for the stages (i) and (ii) also indicate the number of received tracks and trajectories during offloading, respectively.
It can also be seen that every stage, except for stage (v), at least once detects a safety violation that is used to quit offloading.
Consequently, throughout all the simulations, there was no critical situation where an attack was successful.
This result shows that splitting the safety analysis into multiple stages is beneficial.
A more thorough evaluation, if the grid validation stage is actually necessary, is part of our future work.

\begin{table}
\centering
\caption{Usage of different stages of \ours}
\begin{tabular}{l | c | c}
    \hline
    \multirow{2}{*}{Stage}& Number of & Detected \\
    & executions & safety issues \\
    \hline
    (i) Track Validation (Val.) & $1386$ & $24$ \\
    (ii) Trajectory to Map Val. & $582$ & $3$ \\
    (iii) Trajectory Val. with Tracks & $216$ & $15$ \\
    (iv) Trajectory Val. with Detections & $76$ & $1$ \\
    (v) Trajectory Val. with Grid & $188$ & $0$ \\
    \hline
\end{tabular}
\label{table:two-cavs}
\end{table}

\section{Conclusion and Future Work}
In this paper, we have derived an analysis for distributed SOAs that allows different handling of remote and local services.
We have used this analysis to present \ours, a Multi-staged Framework for Safety Analysis that we have integrated into SOFOF as an extension to be able to handle safety-critical offloading scenarios.
Our evaluations show that the integration of \ours into SOFOF is beneficial for a safe usage of function offloading in autonomous driving, especially if data from remote services is erroneous, e.g., manipulated by an external attacker.

In our future work, we want to further investigate possible safety methods and fallback strategies for different autonomous driving services, e.g., using a safe halt mechanism similar to~\cite{ackermann22}.
We also want to integrate SOFOF with \ours in our real-world test vehicle to evaluate the capabilities in real-world traffic scenarios.

\balance
\bibliographystyle{IEEEtran}
{ \footnotesize \bibliography{biblio} }

\end{document}

%% file: img/overview.tex
\tikzstyle{arrow} = [thick,->,>=stealth]
\tikzstyle{doublearrow} = [thick,<->,>=stealth]

\definecolor{darkblue}{HTML}{1f4e79}
\definecolor{salmon}{HTML}{ff9c6b}
\definecolor{maroon}{HTML}{b81414}

\def \minwidth {1.2cm}

\tikzstyle{box} = [draw, rectangle, minimum width=1.9cm, minimum height=0.9cm, font=\footnotesize]
\tikzstyle{rounded box} = [draw, rectangle, rounded corners, minimum height=.4cm, minimum width=\minwidth, font=\scriptsize]
\def \dist {0cm}

\def \colorlocal {custom-blue}
\def \colorremote {muted-red}

\pgfdeclarelayer{background layer}
\pgfdeclarelayer{foreground layer}
\pgfsetlayers{background layer,main,foreground layer}

\begin{tikzpicture}
\begin{pgfonlayer}{background layer}
\node (cav-v2x) [rounded box, top color=\colorlocal!80, bottom color=\colorlocal!60, align=center, xshift=.6cm, yshift=.1cm] {V2X};
\node (det) [rounded box, top color=\colorlocal!80, bottom color=\colorlocal!60, above of=cav-v2x, xshift=-3cm, yshift=1cm] {detection};
\node (mot) [rounded box, top color=\colorlocal!80, bottom color=\colorlocal!60, above of=cav-v2x, xshift=-1.8cm, yshift=0cm] {tracking};
\node (ego) [rounded box, top color=\colorlocal!80, bottom color=\colorlocal!60, above of=cav-v2x, xshift=-2.0cm, yshift=-.8cm] {ego motion};
\node (env) [rounded box, top color=\colorlocal!80, bottom color=\colorlocal!60, opacity=.2, above of=cav-v2x, text width=\minwidth, align=center, xshift=.25cm, yshift=1cm] {environment\\modeling};
\node (map) [rounded box, top color=\colorlocal!80, bottom color=\colorlocal!60, opacity=.2, above of=cav-v2x, text width=\minwidth, align=center, xshift=-1.4cm, yshift=1.1cm] {map provider};
\node (tpl) [rounded box, top color=\colorlocal!80, bottom color=\colorlocal!60, opacity=.2, above of=cav-v2x, text width=\minwidth, align=center,  xshift=1.9cm, yshift=1.1cm] {trajectory\\ planning};
\node (fu-off) [rounded box, top color=\colorlocal!80, bottom color=\colorlocal!60, above of=cav-v2x, text width=\minwidth, align=center, xshift=.4cm, yshift=-.1cm, ] {function\\ offloading};
\node (ctrl) [rounded box, top color=\colorlocal!80, bottom color=\colorlocal!60, above of=cav-v2x,text width=\minwidth, align=center, xshift=2.4cm,yshift=-.3cm] {control};

\node (mec-v2x) [rounded box, top color=\colorremote!80, bottom color=\colorremote!60, below of=cav-v2x] {V2X};
\node (mec-fu-off) [rounded box, top color=\colorremote!80, bottom color=\colorremote!60, text width=\minwidth, align=center, below of=mec-v2x, xshift=-.4cm, yshift=.2cm] {function\\offloading};
\node (mec-tpl) [rounded box, top color=\colorremote!80, bottom color=\colorremote!60, text width=\minwidth, align=center, below of=mec-v2x, xshift=-2.2cm, yshift=.6cm] {trajectory\\planning};
\node (mec-env) [rounded box, top color=\colorremote!80, bottom color=\colorremote!60, text width=\minwidth, align=center, below of=mec-v2x, xshift=-1.6cm, yshift=-.8cm] {environment\\modeling};
\node (mec-map) [rounded box, top color=\colorremote!80, bottom color=\colorremote!60, text width=\minwidth, align=center, left of=mec-env, xshift=-.8cm, yshift=.1cm] {map provider};
\node (mec-det) [rounded box, top color=\colorremote!80, bottom color=\colorremote!60, text width=\minwidth, align=center, below of=mec-v2x, xshift=2.4cm, yshift=.2cm] {detection};
\node (mec-mot) [rounded box, top color=\colorremote!80, bottom color=\colorremote!60, text width=\minwidth, align=center, below of=mec-v2x, xshift=1.8cm, yshift=-.6cm] {tracking};

\node[minimum size=2pt, text=yellow, right of=mec-v2x, xshift=-.6cm, yshift=.5cm] {\LARGE\Lightning}; 
\node[minimum size=2pt, text=yellow, right of=mec-fu-off, xshift=0.4cm, yshift=0cm] {\HUGE\Lightning};

% MEC and CAV icons
\node[left of=cav-v2x, xshift=-2.7cm, yshift=0.65cm, text=muted-blue] (cav) {\normalsize CAV};
\node[below of=cav, minimum size=2pt, yshift=.5cm] (vehicle) {\includegraphics[width=.9cm]{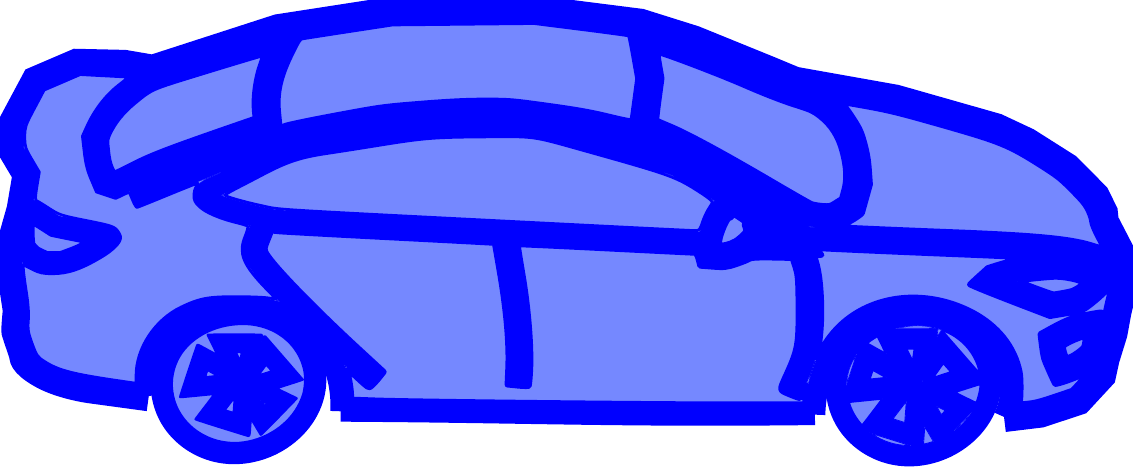}};
\node[left of=mec-v2x, xshift=-2.76cm, yshift=-.8cm, text=muted-red] (mec){\normalsize MEC};
\node[above of=mec, minimum size=2pt, yshift=-.35cm] (server) {\includegraphics[width=.7cm]{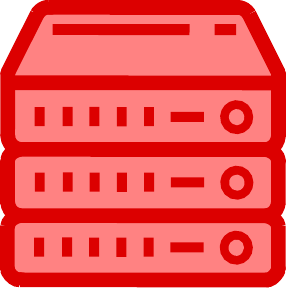}};
\end{pgfonlayer} %% background layer

%% Huge box around cav functions
\node[draw=\colorlocal, rounded corners, fit=(det)(mot)(env)(tpl)(fu-off)(ctrl)(cav-v2x)(vehicle)(cav), inner sep=5pt] (cav-group) {};

%% Huge box around mec functions
\node[draw=\colorremote, rounded corners, fit=(mec-det)(mec-mot)(mec-env)(mec-tpl)(mec-fu-off)(mec-v2x)(mec)(server), inner sep=5pt] (cav-group) {};

\draw [thick, arrow,>=stealth, muted-gray] (det) -- (mot);
\draw [thick, arrow,>=stealth, muted-gold] (mot) -- (fu-off);
\draw [thick, arrow,>=stealth, muted-darkred] (ego) -- (fu-off);
\draw [thick, arrow,>=stealth, muted-darkred] ($(fu-off.south)+(-.2cm,0cm)$) -- ($(cav-v2x.north)+(-.2cm,0cm)$);
\draw [thick, arrow,>=stealth, muted-gold] ($(fu-off.south)+(0cm,0cm)$) -- ($(cav-v2x.north)+(0cm,0cm)$);

\draw [thick, arrow,>=stealth, muted-darkred] ($(cav-v2x.south)+(-.2cm,0cm)$) -- ($(mec-v2x.north)+(-.2cm,0cm)$);
\draw [thick, arrow,>=stealth, muted-gold] ($(cav-v2x.south)+(0cm,0cm)$) -- ($(mec-v2x.north)+(0cm,0cm)$);

\draw [thick, arrow,>=stealth, muted-darkred] ($(mec-v2x.south)+(-.2cm,0cm)$) -- ($(mec-fu-off.north)+(-.2cm,0cm)$);
\draw [thick, arrow,>=stealth, muted-gold] ($(mec-v2x.south)+(0cm,0cm)$) -- ($(mec-fu-off.north)+(0cm,0cm)$);
\draw [thick, arrow,>=stealth, muted-darkred] ($(mec-fu-off.south)+(-.4cm,0)$) -- ($(mec-env.north)+(.3cm,0)$);
\draw [thick, arrow,>=stealth, muted-gold] ($(mec-fu-off.south)+(-.2cm,0)$) -- ($(mec-env.north)+(.5cm,0)$);;
\draw [thick, arrow,>=stealth, muted-purple] (mec-map) -- (mec-env);
\draw [thick, arrow,>=stealth, muted-teal] (mec-env) -- (mec-tpl);
\draw [thick, arrow,>=stealth, muted-green] (mec-tpl) -- (mec-fu-off);
\draw [thick, arrow,>=stealth, muted-green] ($(mec-fu-off.north)+(.2cm,0cm)$) -- ($(mec-v2x.south)+(.2cm,0cm)$);

\draw [thick, arrow,>=stealth, muted-green] ($(mec-v2x.north)+(.2cm,0cm)$) -- ($(cav-v2x.south)+(.2cm,0cm)$);

\draw [thick, arrow,>=stealth, muted-green] ($(cav-v2x.north)+(.2cm,0cm)$) -- ($(fu-off.south)+(.2cm,0cm)$);
\draw [thick, arrow,>=stealth, muted-green] (fu-off) -- (ctrl);

\draw [thick, arrow,>=stealth, muted-gray] (mec-det) -- (mec-mot);
\draw [thick, arrow,>=stealth, muted-gold] (mec-mot) -- (mec-env);

% \draw [thick, arrow,>=stealth, muted-darkred] ($(ctrl.east)$) -- ($(ctrl.east)+(0.8cm,0)$);

%% Create legend for arrows
\begin{pgfonlayer}{foreground layer}
\node[left of=mec-v2x, xshift=3.3cm, yshift=1.22cm] (arrow-help-1-1) {};
\node[right of=arrow-help-1-1, xshift=-.02cm, text=muted-gray] (arrow-help-1-2) {\scriptsize detections};
\draw [thick, arrow,>=stealth, muted-gray] (arrow-help-1-1) -- (arrow-help-1-2);

\node[below of=arrow-help-1-1, yshift=.7cm] (arrow-help-2-1) {};
\node[right of=arrow-help-2-1, xshift=-.22cm, text=muted-gold] (arrow-help-2-2) {\scriptsize tracks};
\draw [thick, arrow,>=stealth, muted-gold] (arrow-help-2-1) -- (arrow-help-2-2);

\node[below of=arrow-help-2-1, yshift=.7cm] (arrow-help-3-1) {};
\node[right of=arrow-help-3-1, xshift=.05cm, text=muted-darkred] (arrow-help-3-2) {\scriptsize ego-motion};
\draw [thick, arrow,>=stealth, muted-darkred] (arrow-help-3-1) -- (arrow-help-3-2);

\node[below of=arrow-help-3-1, yshift=.7cm] (arrow-help-4-1) {};
\node[right of=arrow-help-4-1, xshift=-.5cm, text=muted-purple] (arrow-help-4-2) {};
\node[right of=arrow-help-4-1, xshift=-.3cm, yshift=-.04cm, text=muted-purple] (arrow-help-4-3) {\scriptsize map};
\draw [thick, arrow,>=stealth, muted-purple] (arrow-help-4-1) -- (arrow-help-4-2);

\node[below of=arrow-help-4-1, yshift=.7cm] (arrow-help-5-1) {};
\node[right of=arrow-help-5-1, xshift=-.5cm, text=muted-teal] (arrow-help-5-2) {};
\node[right of=arrow-help-5-1, xshift=.005cm, yshift=.03cm, text=muted-teal] (arrow-help-5-3) {\scriptsize env model};
\draw [thick, arrow,>=stealth, muted-teal] (arrow-help-5-1) -- (arrow-help-5-2);

\node[below of=arrow-help-5-1, yshift=.7cm] (arrow-help-6-1) {};
\node[right of=arrow-help-6-1, xshift=-.06cm, text=muted-green] (arrow-help-6-2) {\scriptsize trajectory};
\draw [thick, arrow,>=stealth, muted-green] (arrow-help-6-1) -- (arrow-help-6-2);

\end{pgfonlayer}

%% Huge box around legend
\begin{pgfonlayer}{main}
\node[draw=black, fill=white, rounded corners, fit=(arrow-help-1-1)(arrow-help-1-2)(arrow-help-5-2)(arrow-help-3-2)(arrow-help-6-2),, inner sep=0pt] (legend-group) {};
\end{pgfonlayer}

\end{tikzpicture}

%% file: img/update_hierarchical_structure.tex
\tikzstyle{arrow} = [thick,->,>=stealth]
\tikzstyle{doublearrow} = [thick,<->,>=stealth]

\definecolor{darkblue}{HTML}{1f4e79}
\definecolor{salmon}{HTML}{ff9c6b}
\definecolor{maroon}{HTML}{b81414}

\tikzstyle{box} = [draw, rectangle, minimum width=1.9cm, minimum height=0.9cm, font=\footnotesize]
\tikzstyle{rounded box} = [draw, rectangle, rounded corners, minimum height=.55cm, minimum width=1.4cm, font=\footnotesize]
\tikzstyle{legend box} = [draw, rectangle, rounded corners, minimum height=.1cm, minimum width=.1cm, font=\footnotesize]

\def \dist {0.2cm}
\def \cavclr {custom-blue!70!white}
\def \mecclr {muted-red!80!white}

\def \colorlocal {custom-blue}
\def \colorremote {muted-red}

\begin{tikzpicture}
\node (root) [rounded box, top color=\colorlocal!80, bottom color=\colorlocal!60] {actuation};
\node (cav-ctrl) [rounded box, below of=root, yshift=\dist, top color=\colorlocal!80, bottom color=\colorlocal!60] {control};
\node (cav-tpl) [rounded box, below of=cav-ctrl, yshift=\dist, top color=\colorlocal!80, bottom color=\colorlocal!60, align=center] {trajectory planning};
\node (mec-tpl) [rounded box, below of=cav-ctrl, xshift=4.5cm, yshift=\dist, top color=\colorremote!80, bottom color=\colorremote!60, align=center] {trajectory planning};
\node (cav-env) [rounded box, below of=cav-tpl, yshift=\dist, top color=\colorlocal!80, bottom color=\colorlocal!60] {environment};
\node (mec-env) [rounded box, below of=mec-tpl, yshift=\dist, top color=\colorremote!80, bottom color=\colorremote!60] {environment};
% \node (cav-fusion) [rounded box, below of=cav-env, xshift=1.6cm, yshift=\dist, top color=\colorlocal!80, bottom color=\colorlocal!60] {fusion};
\node (cav-ego) [rounded box, below of=cav-env, yshift=\dist, top color=\colorlocal!80, bottom color=\colorlocal!60] {ego motion};
\node (cav-map) [rounded box, below of=cav-env, xshift=-1.65cm, yshift=\dist*.5, top color=\colorlocal!80, bottom color=\colorlocal!60, align=center] {map\\provider};
% \node (mec-fusion) [rounded box, below of=mec-env, yshift=\dist, xshift=-.9cm, top color=\colorremote!80, bottom color=\colorremote!60] {fusion};
\node (mec-map) [rounded box, below of=mec-env, yshift=\dist*.5, xshift=0.75cm, top color=\colorremote!80, bottom color=\colorremote!60, align=center] {map\\provider};
\node (cav-track) [rounded box, below of=cav-env, xshift=1.6cm, yshift=\dist, top color=\colorlocal!80, bottom color=\colorlocal!60] {tracking};
\node (mec-track) [rounded box, below of=mec-env, yshift=\dist, xshift=-.9cm, top color=\colorremote!80, bottom color=\colorremote!60] {tracking};
\node (cav-det) [rounded box, below of=cav-track, yshift=\dist, top color=\colorlocal!80, bottom color=\colorlocal!60] {detection};
\node (mec-det) [rounded box, below of=mec-track, yshift=\dist, top color=\colorremote!80, bottom color=\colorremote!60] {detection};

\draw [arrow] (cav-ctrl) -- (root);
\draw [arrow] (cav-tpl) -- (cav-ctrl);
\draw [arrow] (mec-tpl) -- (cav-ctrl);
\draw [arrow] (cav-env) -- (cav-tpl);
\draw [arrow] (mec-env) -- (mec-tpl);
\draw [arrow] (cav-ego) -- (cav-env);
\draw [arrow] ($(cav-ego.north east)+(-.1cm,0)$) -- (mec-env);
\draw [arrow] ($(cav-map.north east)+(-.1cm,0)$) -- (cav-env);
\draw [arrow] (mec-map) -- (mec-env);

% \draw [arrow] (cav-track) -- (cav-fusion);
% \draw [arrow] (mec-track) -- (cav-fusion);
% \draw [arrow] (mec-track) -- (mec-fusion);
% \draw [arrow] (cav-track) -- (mec-fusion);
% \draw [doublearrow] (cav-fusion) -- (mec-fusion);

\draw [arrow] (cav-track) -- (cav-env);
\draw [arrow] (mec-track) -- (cav-env);
\draw [arrow] (mec-track) -- (mec-env);
\draw [arrow] (cav-track) -- (mec-env);

\draw [arrow] (cav-det) -- (cav-track);
\draw [arrow] (mec-det) -- (cav-track);
\draw [arrow] (mec-det) -- (mec-track);
\draw [arrow] (cav-det) -- (mec-track);

%% Create Legend
\node (legend-cav) [legend box, right of=root, xshift=1.2cm, yshift=-.1cm, top color=\colorlocal!80, bottom color=\colorlocal!60] {};
\node (legend-cav-txt) [right of=legend-cav, xshift=-.5cm, font=\footnotesize] {CAV};
\node (legend-mec) [legend box, right of=legend-cav, xshift=.2cm, top color=\colorremote!80, bottom color=\colorremote!60] {};
\node (legend-mec-txt) [right of=legend-mec, xshift=-.5cm, font=\footnotesize] {MEC};
%% Box around legend
\node[draw=black, rounded corners, fit=(legend-cav)(legend-cav-txt)(legend-mec)(legend-mec-txt), inner sep=2pt] (cav-group) {};
    
\end{tikzpicture}

%% file: img/framework.tex
\begin{tikzpicture}

\def \minwidth {0.9cm}
\def \minheight {0.3cm}
\def \customxshift {-.25cm}
\def \lwidth {1pt}
\def \verticalshift {-.75cm}

\def \colorlocal {muted-blue}
\def \colorremote {muted-red}

\tikzstyle{rounded box} = [draw, rectangle, rounded corners, minimum height=.6cm, minimum width=\minwidth, font=\scriptsize]

\tikzstyle{remote} = [draw, top color=\colorremote, bottom color=\colorremote!80, shape=motshape, minimum width=\minwidth, minimum height=\minheight, text=black, line width=\lwidth, font=\footnotesize];
\tikzstyle{local} = [draw, top color=\colorlocal!80, bottom color=\colorlocal!60, shape=motshape, minimum width=\minwidth, minimum height=\minheight, text=black, line width=\lwidth, font=\footnotesize];

% Top group 1
\node[remote] (mot1) {MOT};
\node[local, right of=mot1, xshift=\customxshift] (env1) {ENV};
\node[local, right of=env1, xshift=\customxshift] (tpl1) {TPL};

% Top group 2
\node[remote, right of=tpl1, xshift=.3cm] (mot2) {MOT};
\node[remote, right of=mot2, xshift=\customxshift] (env2) {ENV};
\node[remote, right of=env2, xshift=\customxshift] (tpl2) {TPL};

% Top group 3
\node[local, right of=tpl2, xshift=.3cm] (mot3) {MOT};
\node[remote, right of=mot3, xshift=\customxshift] (env3) {ENV};
\node[remote, right of=env3, xshift=\customxshift] (tpl3) {TPL};

\pgfdeclarehorizontalshading{goldtogreen}{100bp}{
  color(0bp)=(muted-gold!70!white);
  color(49bp)=(muted-gold!70!white);  % ← stay red until 70% across
  color(51bp)=(muted-green!70!white);  % ← stay red until 70% across
  color(100bp)=(muted-green!70!white)
}

\node[rounded box, below of=mot1, minimum width=5.6cm, xshift=2.15cm, yshift=.2cm, font=\small, fill=muted-green!70!white] (mot-val) {Track Validation};
\node[rounded box, below of=mot1, minimum width=8.4cm, xshift=3.55cm, yshift=\verticalshift+.2cm, font=\small, fill=muted-green!70!white] (map-val) {Trajectory to Map Validation};
\node[rounded box, below of=mot1, minimum width=5.6cm, xshift=4.95cm, yshift=2*\verticalshift+.2cm, font=\small, shading=goldtogreen, shading angle=0] (track-val) {Trajectory Validation with Tracks};
\node[rounded box, below of=mot1, minimum width=8.4cm, xshift=3.55cm, yshift=3*\verticalshift+.2cm, font=\small, fill=muted-gold!70!white] (det-val) {Trajectory Validation with Detections};
\node[rounded box, below of=mot1, minimum width=5.6cm, xshift=2.15cm, yshift=4*\verticalshift+.2cm, font=\small,shading=goldtogreen, shading angle=180] (grid-val) {Trajectory Validation with Grid Map};

% arrows 1st to 2nd

\def \arrowoff {-.15cm}
\draw [->] ($(mot-val.south west)+ (.4cm,0)$) -- ($(mot-val.south west)+(.4cm,\arrowoff)$);
\draw [->] ($(mot-val.south)+ (-.4cm,0)$) -- ($(mot-val.south)+(-.4cm,\arrowoff)$);
\draw [->] ($(mot-val.south)+ (.4cm,0)$) -- ($(mot-val.south)+(.4cm,\arrowoff)$);
\draw [->] ($(mot-val.south east)+ (-.4cm,0)$) -- ($(mot-val.south east)+(-.4cm,\arrowoff)$);

% arrows 2nd to 3rd
\draw [->] ($(map-val.south)+ (-1cm,0)$) -- ($(map-val.south)+(-1cm,\arrowoff)$);
\draw [->] ($(map-val.south)+ (1cm,0)$) -- ($(map-val.south)+(1cm,\arrowoff)$);
\draw [->] ($(map-val.south)+ (1.8cm,0)$) -- ($(map-val.south)+(1.8cm,\arrowoff)$);
\draw [->] ($(map-val.south east)+ (-.4cm,0)$) -- ($(map-val.south east)+(-.4cm,\arrowoff)$);

% arrows 2nd to 4th
\draw [->] ($(map-val.south west)+ (.4cm,0)$) -- ($(det-val.north west)+(.4cm,0)$);
\draw [->] ($(map-val.south)+ (-1.8cm,0)$) -- ($(det-val.north)+(-1.8cm,0cm)$);

% arrows 3rd to 4th
\draw [->] ($(track-val.south west)+ (.4cm,0)$) -- ($(track-val.south west)+(.4cm,\arrowoff)$);
\draw [->] ($(track-val.south)+ (-.4cm,0)$) -- ($(track-val.south)+(-.4cm,\arrowoff)$);
\draw [->] ($(track-val.south)+ (.4cm,0)$) -- ($(track-val.south)+(.4cm,\arrowoff)$);
\draw [->] ($(track-val.south east)+ (-.4cm,0)$) -- ($(track-val.south east)+(-.4cm,\arrowoff)$);

% arrows 4th to 5th
\draw [->] ($(det-val.south west)+ (.4cm,0)$) -- ($(det-val.south west)+(.4cm,\arrowoff)$);
\draw [->] ($(det-val.south)+ (-1.8cm,0)$) -- ($(det-val.south)+(-1.8cm,\arrowoff)$);
\draw [->] ($(det-val.south)+ (-1cm,0)$) -- ($(det-val.south)+(-1cm,\arrowoff)$);
\draw [->] ($(det-val.south)+ (1cm,0)$) -- ($(det-val.south)+(1cm,\arrowoff)$);

% Dashed lines
\node (help1) at ($(tpl1)!0.5!(mot2)+(0,6*\verticalshift+.3cm)$) {};
\node (help2) at ($(tpl3.east)+(.2cm,6*\verticalshift+.3cm)$) {};

\begin{scope}[on background layer]
    \fill[black!15] ($(mot1.west)+(-.2cm,.4cm)$) rectangle (help1);
    \fill[black!5] (help1) rectangle ($(tpl2)!0.5!(mot3)+(0,.4cm)$);
    \fill[black!15] (help2) rectangle ($(tpl2)!0.5!(mot3)+(0,.4cm)$);
\end{scope}

% \draw [thick, muted-gray, dashed] ($(c1)+(0,\minheight/2)$) -- ($(c1)+(0,-4.9cm)$);
% \draw [thick, muted-gray, dashed] ($(c2)+(0,\minheight/2)$) -- ($(c2)+(0,-4.9cm)$);

\end{tikzpicture}